\documentclass[preprint,12pt]{elsarticle}
\biboptions{numbers,sort&compress}

\usepackage{amssymb}
\usepackage{amsmath}
\usepackage{caption}
\usepackage{algorithm}
\usepackage{algpseudocode}
\usepackage{booktabs} 
\usepackage{multirow}

\begin{document}

\begin{frontmatter}



\title{DiffMark: Diffusion-based Robust Watermark Against Deepfakes} 
\tnotetext[funding]{Funding: This work was supported by the National Natural Science Foundation of China [grant numbers 62462060, 62302427]; the Natural Science Foundation of Xinjiang Uygur Autonomous Region [grant number 2023D01C175]; the Tianshan Talent Training Program [grant number 2022TSYCLJ0036].}
\tnotetext[license]{© 2025. This manuscript version is made available under the CC BY-NC-ND 4.0 license https://creativecommons.org/licenses/by-nc-nd/4.0/. Link to formal publication: https://doi.org/10.1016/j.inffus.2025.103801.}

\author[1]{Chen Sun} 
\ead{sunc@stu.xju.edu.cn}

\author[1]{Haiyang Sun} 

\author[1,2]{Zhiqing Guo\corref{cor}}
\ead{guozhiqing@xju.edu.cn}

\author[3]{Yunfeng Diao}

\author[1]{Liejun Wang}

\author[1]{Dan Ma\corref{cor}}

\ead{madan@xju.edu.cn}

\author[4]{Gaobo Yang}

\author[5]{Keqin Li}

\cortext[cor]{Corresponding author}

\affiliation[1]{organization={College of Computer Science and Technology, Xinjiang University},
	city={Urumqi},
	country={China}}

\affiliation[2]{organization={Silk Road Multilingual Cognitive Computing International Cooperation Joint Laboratory},
	city={Urumqi},
	country={China}}

\affiliation[3]{organization={School of Computer Science and Information Engineering, Hefei University of Technology},
	city={Hefei},
	country={China}}

\affiliation[4]{organization={College of Computer Science and Electronic Engineering, Hunan University},
	city={Changsha},
	country={China}}

\affiliation[5]{organization={Department of Computer Science, State University of New York},
	city={New York},
	country={USA}}

\begin{abstract}
Deepfakes pose significant security and privacy threats through malicious facial manipulations. 
While robust watermarking can aid in authenticity verification and source tracking, existing methods often lack sufficient robustness against Deepfake manipulations.
Diffusion models have demonstrated remarkable performance in image generation, enabling the seamless fusion of watermark with image during generation.
In this study, we propose a novel robust watermarking framework based on diffusion model, called DiffMark. By modifying the training and sampling scheme, we take the facial image and watermark as conditions to guide the diffusion model to progressively denoise and generate the corresponding watermarked image. 
In the construction of facial condition, we weight the facial image by a timestep-dependent factor that gradually reduces the guidance intensity with the decrease of noise, thus better adapting to the sampling process of diffusion model.
To achieve the fusion of watermark condition, we introduce a cross information fusion (CIF) module that leverages a learnable embedding table to adaptively extract watermark features and integrates them with image features via cross-attention.
To enhance the robustness of the watermark against Deepfake manipulations, we integrate a frozen autoencoder during training phase to simulate Deepfake manipulations. Additionally, we introduce Deepfake-resistant guidance that employs specific Deepfake model to adversarially guide the diffusion sampling process to generate more robust watermarked images.
Experimental results demonstrate the effectiveness of the proposed DiffMark on typical Deepfakes.
Our code will be available at https://github.com/vpsg-research/DiffMark.
\end{abstract}




\end{frontmatter}



\section{Introduction}\label{}

In recent years, the remarkable development of generative models has significantly propelled the advancement of Deepfake \cite{chen2020simswap,choi2018stargan,rochow2024fsrt,LI2024102456}. Deepfake has shown vast potential for applications in industries such as film production and advertising. 
However, its malicious use has brought about significant security risks associated with face forgery and profound ethical concerns. Malicious Deepfakes severely threaten personal privacy and social stability, facilitate the spread of disinformation, undermine trust in digital media and institutions. To address both the security challenges and ethical risks posed by malicious face forgery, developing appropriate countermeasures is becoming increasingly crucial and urgent.

\begin{figure}[!t]
	\centering
	\includegraphics[width=\textwidth]{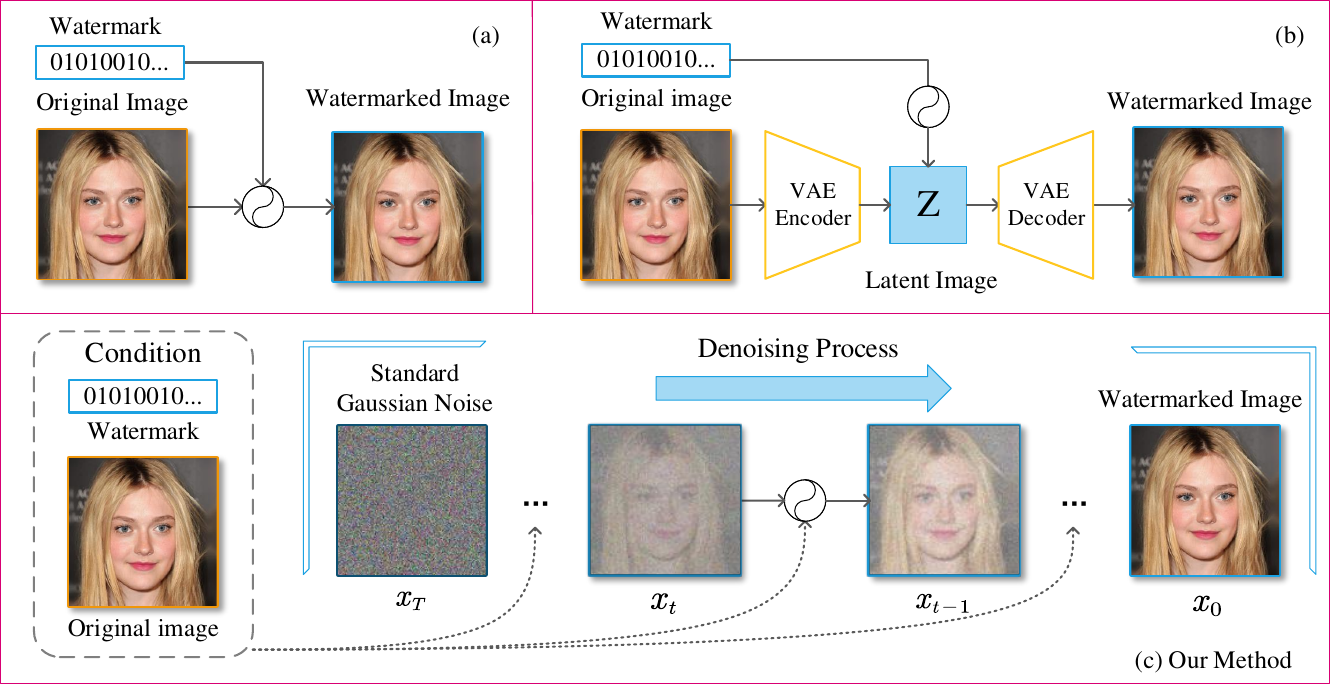}
	\caption{The difference between our method and the existing methods: (a) Traditional pixel-space methods directly embed the watermark in the pixel space of the image; (b) Latent-space methods that transform image into latent representation for watermark embedding; (c) Our method initiates from standard Gaussian distribution, using facial image and watermark as conditions to guide the diffusion model denoising for watermarked image generation.}
	\label{fig:first}
\end{figure}

Passive forensics \cite{guo2023ldfnet,QIU2025103087,TOLOSANA2020131,YOON2024102424} primarily determines the authenticity of facial images by analyzing the subtle traces or artifacts, which is essentially a binary classification task. 
As they only operate after the forgery has occurred, they are unable to provide reliable traceability. 
Proactive forensics \citep{wang2021faketagger,wu2023sepmark,wang2023robust,wang2024lampmark} has the advantage of preemption, with most methods employing deep watermarking for authenticity verification and  source tracking. The fundamental principle is to embed watermarks into facial images before they are released. 
These embedded watermarks are visually imperceptible and mostly can be robustly extracted after Deepfake manipulations for source tracing. 
Although the function of embedded watermarks may not be limited to traceability, we believe that robust traceability is fundamental and almost indispensable. 
However, existing deep watermarking methods often fall short in robustness when confronted with diverse Deepfake manipulations. 
As illustrated in Fig.~\ref{fig:first}, they can be broadly classified into two categories. The conventional pixel-space methods \cite{jia2021mbrs,huang2023arwgan} typically employ a neural network to directly embed watermarks into images in pixel space. Although the perturbation induced by watermark is small, 
it often lacks enough robustness against various attacks.
The latent-space methods \cite{Bui_2023_CVPR,meng2025latent} transform images into latent representations for watermark embedding, which improves the robustness of the watermark. However, they are prone to cause the reconstructed image to lose image details. 

To address the limitations of existing methods in terms of robustness against Deepfake manipulations, while maintaining image quality, we propose a novel diffusion-based robust watermarking framework, called DiffMark. Diffusion models \citep{ho2020denoising,song2020denoising,dhariwal2021diffusion,rombach2022high} have demonstrated remarkable performance in image generation. 
It is possible to employ the diffusion model to generate watermarked images.
However, sampling from a standard Gaussian distribution introduces significant randomness, which is ideal for diverse styles of image generation but unsuitable for watermark embedding in deterministic image. To mitigate this, we construct both facial image and watermark as conditions to guide the denoising process, ensuring the generation of corresponding watermarked image. 
To enhance the robustness of watermark against Deepfakes, we adopt a two-stage strategy. In the training phase, we incorporate a frozen VQGAN \cite{esser2021taming} autoencoder to simulate Deepfake manipulations, achieving comparable robustness against Deepfakes as well as common distortions such as jpeg compression. In the inference phase, unlike the existing methods that perform fusion only once, we take the facial image and watermark as diffusion conditions and fuse them with the t-step noisy image for several timesteps during the sampling process.
Inspired by classifier guidance \cite{dhariwal2021diffusion}, we propose the Deepfake-resistant guidance that incorporates the specific Deepfake model into the sampling process. By using the gradient of watermark extraction after Deepfake manipulations to guide the sampling process, the diffusion model can generate more robust watermarked images. 
The Deepfake-resistant guidance can be viewed as a training-free enhancement module, offering greater flexibility.

For the construction of the facial condition, we do not directly use the facial image. Instead, we apply the timestep-dependent coefficient of the noise term as a scaling factor to the facial image. This approach aims to maintain strong semantic guidance during the early stages of sampling to ensure the direction of generation. As the denoising timestep progresses and the noise intensity decreases, the influence of the facial condition correspondingly diminishes, achieving progressive conditional guidance that better adapts to the sampling process of the diffusion model. For the fusion of watermark condition, we design a cross information fusion (CIF) module. It is noticed that the widely used binary watermark includes position and bit values. Thus, we combine these two kinds of information to construct unique embedding indices, enabling adaptive watermark feature extraction via an optimized embedding table lookup mechanism. The extracted watermark features are then deeply integrated with facial features through cross-attention. 

In summary, our contributions are as follows:
\begin{itemize}
	\item We propose DiffMark, a novel diffusion-based robust watermark framework. Innovatively, we construct facial image and watermark as conditions to guide the diffusion model to gradually denoise and generate the corresponding watermarked image. 
	\item To enhance the robustness of watermark against Deepfakes, we incorporate a pre-trained frozen autoencoder to simulate Deepfake manipulations during training and introduce Deepfake-resistant guidance during the diffusion model's sampling process.
	\item For watermark condition fusion, we design a cross information fusion module that employs positional-bit encoding to generate embedding indices for watermark feature retrieval, enabling cross-attention-driven integration with facial features.
\end{itemize}

\section{RELATED WORKS}

\subsection{Deep Robust Watermarking}

In recent years, robust watermarking based on deep learning has garnered extensive research attention. Researchers have proposed various methods to enhance the robustness and imperceptibility of watermark. 
\citet{zhu2018hidden} proposed the first end-to-end trainable framework based on deep neural networks for robust watermark hiding in images. 
\citet{jia2021mbrs} introduced a novel training method using mini-batch of real and simulated JPEG compression to enhance the JPEG robustness of watermark.  
\citet{ma2022towards} combined invertible and non-invertible mechanisms to enhance the imperceptibility and robustness of blind watermarking against various noises. 
\citet{huang2023arwgan} introduced a GAN-based attention-guided robust image watermarking method, which highlights essential features for better integrating image and watermark features. 
\citet{tan2024waterdiff} proposed WaterDiff, which utilized a pretrained diffusion-based autoencoder for reversible mapping and image watermarking via discrete wavelet transform (DWT). However,  the reversible mapping of diffusion-based autoencoder tends to lose image details. 
In contrast, our method integrates image and watermark throughout the diffusion sampling process, thereby preserving more image detail.

To prevent the malicious Deepfakes, researchers realize the value of watermarking technology and employ deep watermarking for Deepfake proactive forensics.
\citet{wang2021faketagger} embedded messages called tags in facial images, which can be recovered after various Deepfake manipulations for source tracing.
\citet{wu2023sepmark} introduced the deep separable watermarking framework, utilizing two decoders operating under different robustness levels to simultaneously achieve source tracing and Deepfake detection. 
\citet{wang2023robust} assigned the facial identity semantics to watermarks, integrates a chaotic encryption system for watermark confidentiality, enabling proactive detection and source tracing against face swapping. 
\citet{zhang2024editguard} embedded dual invisible watermarks into original images, not only protecting image copyrights but also locating tampered regions. 
\citet{ijcai2024p673} fine-tuned robust watermarking into adversarial watermarking, enhancing the detectability of passive Deepfake detector while maintaining the traceability.
\citet{zhang2024dual} proposed a novel traceable adversarial watermark method, which can simultaneously track face copyrights and disrupt the face swapping model.
\citet{wang2024lampmark} leveraged the structure-sensitive properties of facial landmarks to create binary landmark perceptual watermarks for Deepfake proactive forensics. 

The existing methods usually fuse images and watermarks in a single step through neural networks.
In contrast, we take the image and watermark as diffusion conditions, iteratively fusing them with the noisy image at each timestep during the sampling process, enhancing the robustness of watermark. 
We further introduce Deepfake-resistant guidance to guide the sampling process to generate more robust watermarked image against Deepfake manipulations to some extent. 

\subsection{Diffusion Model}

Diffusion models, as an emerging class of generative models, have gradually garnered significant attention due to their powerful image generation capabilities and theoretical advantages.  \citet{ho2020denoising} introduced DDPM, which iteratively adds noise to data and then learns to reverse the process, achieving high-quality sample generation. \citet{song2020denoising} introduced DDIM, which enables faster sampling without sacrificing generation quality by using a non-Markovian, deterministic sampling process. \citet{dhariwal2021diffusion} proposed classifier guidance, achieving higher sample quality and more controllable generation process. \citet{nichol2021glide} explored text-conditional diffusion model using CLIP guidance and classifier free guidance. \citet{rombach2022high} introduced the Latent Diffusion Model (LDM), which reduces computational costs for high-resolution image generation by operating in latent space while enabling more flexible image generation through various conditions. \citet{zhang2023adding} propose ControlNet, which can add spatial conditions to control the pretrained text-to-image diffusion models.

While image watermarking based on diffusion models is still in its nascent stages and has yet to be fully explored, the powerful image generation capabilities of diffusion models, combined with controllable sampling via conditional inputs, suggest significant potential for diffusion models to serve as a robust watermarking framework.

\begin{figure*}[!t]
	\centering
	\includegraphics[width=\textwidth]{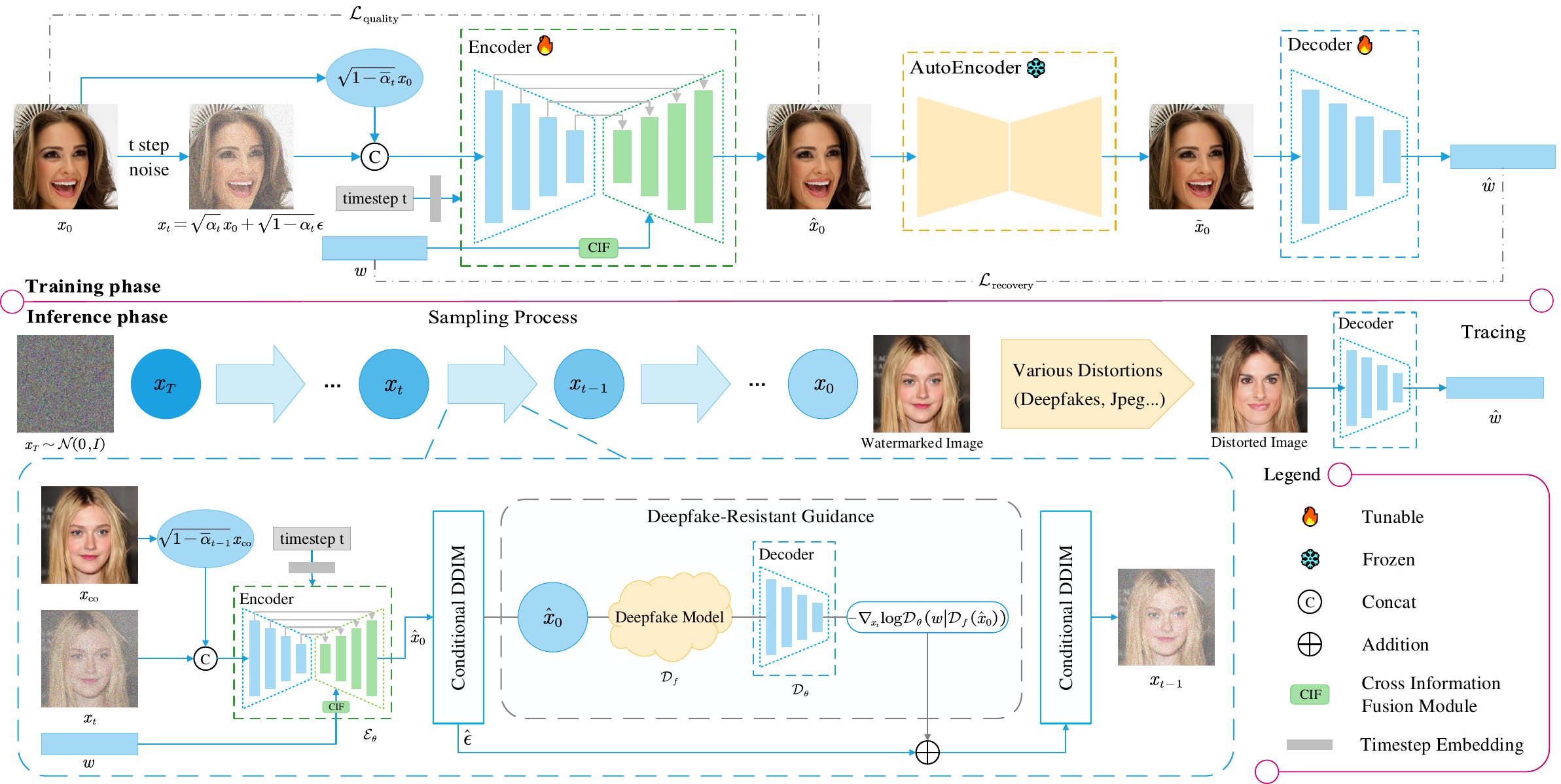}
	\caption{Illustration of the proposed DiffMark. (a) Training Phase: The t-step noised image $x_t$, dynamically scaled facial image $\sqrt{1 - \overline{\alpha}_t} x_0$, watermark $w$ and timestep $t$ are fed into the diffusion encoder to predict the watermarked image $\hat{x}_0$, which is then reconstructed by a frozen autoencoder to produce $\widetilde{x}_0$. The watermark decoder extracts the watermark from $\widetilde{x}_0$. 
	(b) Inference Phase: Initialized with standard Gaussian distribution $x_T$, the Deepfake-resistant guided DDIM sampling process take the scaled facial image $x_c$ and watermark $w$ as conditions to gradually denoise and generate the watermarked image. The watermark can be extracted by the watermark decoder from the distorted image for source tracing.}
	\label{fig:framework}
\end{figure*}

\section{METHODOLOGY}

This section provides a brief introduction to diffusion models, followed by a detailed explanation of our proposed DiffMark. We begin by describing the training phase, where facial images and watermarks are constructed as diffusion conditions to adapt the diffusion model for image watermarking. Next, we detail the inference phase, mainly covering the Deepfake-resistant guided diffusion sampling guidance for the generation of watermarked image. We then introduce the cross information fusion module designed to integrate image features and watermark. Finally, we present the design of the loss function.

\subsection{Preliminaries: Diffusion Models}

Diffusion models are a class of generative models based on iterative noising and denoising. They operate by gradually corrupting the original image $x_0$ with Gaussian noise $\epsilon$ in the forward process until it becomes pure noise $x_T$, then learning to reverse this degradation through a neural network such as U-Net to reconstruct the original image from the noisy image $x_T$ step by step.

The forward process adds noise step by step through a Markov chain, with the mathematical form:  
\begin{equation}
	q(x_t | x_{t-1}) = \mathcal{N}\left(x_t; \sqrt{1-\beta_t} x_{t-1}, \beta_t \mathrm{I}\right)  
\end{equation}
where \(\beta_t \in (0,1)\) controls the noise intensity. 

With the reparameterization trick, \( x_{t} \) at any timestep can be directly computed from \( x_0 \):  
\begin{equation}
	x_t = \sqrt{\bar{\alpha}_t} x_0 + \sqrt{1-\bar{\alpha}_t} \epsilon, \quad \epsilon \sim \mathcal{N}(0, \mathrm{I}) \label{eq:eq2}
\end{equation}
Here, \(\bar{\alpha}_t = \prod_{i=1}^t (1-\beta_i)\), representing the cumulative image degradation.

The reverse process iteratively denoises to restore the original data. It can be described by parameterized Gaussian distribution:  
\begin{equation}
	p_\theta(x_{t-1} | x_t) = \mathcal{N}\left(x_{t-1}; \mu_\theta(x_t, t), \Sigma_\theta(x_t, t)\right)
\end{equation}
where \( \mu_\theta(x_t, t) \) and \( \Sigma_\theta(x_t, t) \) are the mean and variance that can be predicted by a neural network.

\citet{song2020denoising} further proposed Denoising Diffusion Implicit Model (DDIM), which constructs a reverse process of non-Markov chain and allows accelerated sampling:
\begin{equation}
	x_{t-1} = \sqrt{\bar{\alpha}_{t-1}} \left( \frac{x_t - \sqrt{1 - \bar{\alpha}_t} \epsilon_\theta^{(t)}(x_t)}{\sqrt{\bar{\alpha}_t}} \right) + \sqrt{1 - \bar{\alpha}_{t-1}} \epsilon_\theta^{(t)}(x_t) \label{eq:ddim}
\end{equation}
where $\epsilon_\theta^{(t)}(x_t)$ represents the noise predicted by a neural network at timestep $t$.

Moreover, \citet{dhariwal2021diffusion} proposed classifier guidance that employs the gradient of a classifier $p_\phi(y \mid x_t)$ to affect the $t$-step predicted noise $\epsilon_\theta(x_t)$ to steer image generation toward a desired category:
\begin{equation}
	\hat{\epsilon}_t \leftarrow \epsilon_\theta(x_t)\;-\;s\,\sqrt{1 - \bar{\alpha}_t}\;\nabla_{x_t}\log\,p_\phi\bigl(y \mid x_t\bigr)
\end{equation}
where $s$ is a scaling constant that controls the strength of the guidance.

Our DiffMark adopts the DDIM sampler and constructs facial image and watermark as diffusion conditions to ensure the generation of the corresponding watermarked image. 
We further introduce Deepfake-resistant guidance during the sampling process to generate more robust watermarked image against Deepfake manipulations.

\subsection{Diffusion Model with Facial and Watermark Conditions}

The diffusion process of DiffMark is conducted in the pixel space rather than the latent space, as we suppose that the mapping from pixel space to latent space tends to discard image details and preserve only semantic consistency, which is detrimental to the imperceptibility of watermark. Furthermore, we adopt the U-Net as the backbone of the diffusion model, which we refer to as the diffusion encoder (as shown in the encoder of Fig.~\ref{fig:framework}).

During the training phase, we modify the standard diffusion training pipeline to accommodate image watermarking task. Conventional diffusion models primarily take the noise-corrupted image \(x_t\) (obtained by adding \(t\)-step noise to the original image \(x_0\)) and the timestep \(t\) as inputs. They train the diffusion encoder to predict the noise added at timestep \(t\), enabling iterative denoising for image generation. 
For image watermarking, it is evident that the watermark needs to serve as an input to the diffusion encoder. To achieve the fusion of watermark condition, we introduce a cross information fusion module on the two levels of the diffusion encoder's feature hierarchy (detailed in Section~\ref{subsec:CIF}).
\begin{algorithm}[H]
	\caption{DiffMark Training Framework.}
	\label{alg:training}
	\begin{algorithmic}[1]
		\While{not converged}
		\State $x_0 \sim q(x_0)$
		\State $t \sim \text{Uniform}(\{1, \dots, T\})$
		\State $\epsilon \sim \mathcal{N}(0, \mathrm{I})$
		\State $x_t \sim \mathcal{N}(\sqrt{\bar{\alpha}_t}x_0, (1 - \bar{\alpha}_t)\mathrm{I})$
		\State $x_c \leftarrow \sqrt{1 - \bar{\alpha}_t}x_0$
		\State $w \sim \{0, 1\}^L$
		\State $\hat{x}_0 \leftarrow \mathcal{E}_\theta(x_t, t, x_c, w)$
		\State $\hat{w} \leftarrow \mathcal{D}_\theta(AE(\hat{x}_0))$ 
		\State Take gradient descent step on
		\State \quad $\nabla_\theta (\|x_0 - \hat{x}_0\|_2^2 + \alpha \mathcal{L}_\text{lpips}(\hat{x}_0, x_0) + \beta \mathcal{L}_\text{ce}(\hat{w}, w))$
		\EndWhile
	\end{algorithmic}
\end{algorithm}
Nevertheless, training the diffusion encoder by merely incorporating watermark as conditional input introduces considerable randomness during the sample process when initialized from the standard Gaussian distribution——a characteristic beneficial for image generation with various styles but unsuitable for generating specific watermarked image. To reduce the randomness, we dynamically scale the original image \(x_0\) using the coefficient \(\sqrt{1 - \bar{\alpha}_t}\) of the noise \( \epsilon \) and incorporate the scaled image as another conditional input to the diffusion encoder. 
As the timestep \( t \) increases in the diffusion process, the noise coefficient \(\sqrt{1 - \bar{\alpha}_t}\) progressively amplifies, indicating stronger noise intensity and greater uncertainty. To counterbalance this increased randomness, the facial condition \(x_c = \sqrt{1-\bar{\alpha}_{t}}x_0\) is scaled closer to \( x_0 \) at corresponding timesteps. 
For the fusion of facial condition \( x_c \), we concatenate it with the \( t \)-step noised image \( x_t \) in the channel dimension. 
The combination of escalating noise and enhanced conditioning narrows the output distribution of the diffusion sampling process, making it ideal for deterministic image watermarking.

The conventional diffusion models train the diffusion encoder to predict the additive noise $\epsilon$ at each timestep. However, since we dynamically scale the original image \(x_0\) by the noise schedule coefficient and provide it as a conditional input to the diffusion encoder, directly predicting the original image \(x_0\) rather than the noise $\epsilon$ leads to more stable training and faster convergence. For the image watermarking task, the diffusion encoder is expected to predict the watermarked image \(\hat{x}_0\), which represents the original image \(x_0\) containing the imperceptible watermark \(w\). 
However, due to the unknown prior distribution of the watermarked image, we introduce a specialized watermark decoder to extract the watermark and assist the diffusion encoder in predicting the watermarked image \(\hat{x}_0\).
The watermark decoder simply utilize the downsampling trunk of the the diffusion encoder with an output layer at the 8x8 layer to produce the final output.
 
To enhance the robustness of the watermark against Deepfake manipulations, 
we incorporate a pre-trained frozen VQGAN \cite{esser2021taming} autoencoder to simulate the process of image reconstruction in most Deepfake models.
This approach is not only effective against Deepfake manipulations but also provides extra robustness against other common distortions, such as resize and jpeg compression, which is unexpected. This way, the diffusion encoder, watermark decoder, and pre-trained frozen autoencoder collectively constitute the end-to-end training framework of DiffMark.

In summary, the training phase (illustrated in Fig.~\ref{fig:framework} and Algorithm~\ref{alg:training}) begins with the diffusion encoder $\mathcal{E}_\theta$ predicting the watermarked image $\hat{x}_0$ based on the noisy image $x_t$, the dynamically scaled image $x_c$, the watermark $w$, and the timestep $t$. To improve robustness against Deepfakes, a pre-trained frozen autoencoder $AE $ then distorts the watermarked image $x_0$ into $\widetilde{x}_0$. Finally, the watermark decoder $\mathcal{D}_\theta$ extracts the embedded watermark from $\widetilde{x}_0$ and provides feedback to optimize the diffusion encoder's prediction of \( \hat{x}_0 \).

Although our DiffMark focuses on Deepfake proactive forensics of facial images, it may potentially be extended to other image domains. In such cases, the facial image condition could be replaced by any target image without modifying the network architecture, while retraining on the corresponding dataset would be necessary to adapt the framework. 
It should be noted, however, that the Deepfake-resistant guidance elaborated in the following subsection is specific to face forgery and would have to be omitted or replaced when applying DiffMark to non-facial domains.

\subsection{Deepfake-Resistant Guided Diffusion Sampling}
The inference phase of our DiffMark mainly comprises two aspects: watermark embedding and extraction. 
In terms of watermark embedding, unlike traditional deep learning-based watermarking methods that directly embed the watermark into image, we take the facial image and watermark as diffusion conditions and leverage the DDIM sampler \cite{song2020denoising} to progressively denoise and generate the target watermarked image.

It is easy to notice that the sampling process from \(x_T\) to \(x_0\) includes many steps. The diffusion encoder trained through the end-to-end training framework will be utilized in DDIM sampling to facilitate the transition from \(x_t\) to \(x_{t-1}\).  
It is observed that the scaling coefficient applied to the original image \(x_0\) matches the coefficient of the added Gaussian noise \( \epsilon \) during training phase. 
Therefore, it is important to find the noise term to determine the scaling factor of the image during the diffusion sampling process. For DDIM sampling, 
we notice the second term in the Equation \ref{eq:ddim} reintroduces the predicted noise, so the coefficient \( \sqrt{1-\bar{\alpha}_{t-1}} \) naturally serves as the scaling factor for the facial condition, yielding $x_c = \sqrt{1-\bar{\alpha}_{t-1}}x_\text{co}$.

This strategic design of facial condition not only prevents the diffusion encoder from developing an over-reliance on the original image \(x_\text{co}\), but also ensures proper attention to the \( t \)-step noised image \(x_t\). It is the premise of Deepfake-resistant guidance, which requires calculating the gradient with respect to \(x_t\). This approach tightly couples watermarked image generation with the whole sampling process of diffusion model.
It is described in the previous section that we freeze a pre-trained autoencoder in the training phase to enhance the robustness of the watermark against various distortions. To further enhance the robustness of watermark against Deepfakes, we proposed the Deepfake-resistant guidance during the DDIM sampling process. 

\begin{algorithm}[H]
	\caption{DDIM Sampling with Deepfake-Resistant Guidance.}\label{alg:sampling}
	\begin{algorithmic}[1]
		\State \textbf{Input:} facial image $x_\text{co}$, watermark $w\sim\{0, 1\}^L$, boolean $cond$, gradient scale $s$
		\State \textbf{Output:} watermarked image $x_0$ corresponding to $x_\text{co}$
		\State $x_T \sim \mathcal{N}(0, \mathrm{I})$
		\For{$t$ from $T$ to $1$}
		\State $x_c \leftarrow \sqrt{1-\bar{\alpha}_{t-1}}x_\text{co}$
		\State $\hat{x}_0 \leftarrow \mathcal{E}_\theta(x_t, t, x_c, w)$
		\State $\hat{\epsilon} \leftarrow \frac{1}{\sqrt{1 - \bar{\alpha}_t}} x_t - \sqrt{\frac{\bar{\alpha}_t}{1 - \bar{\alpha}_t}} \hat{x}_0$
		\If{$cond$}
		\State $\hat{\epsilon} \leftarrow \hat{\epsilon} - s\,\sqrt{1 - \bar{\alpha}_t} \nabla _{x_t}\log \mathcal{D}_\theta\left( w|\mathcal{D}_f(\hat{x}_0) \right) $
		\EndIf
		\State $x_{t-1} \leftarrow \sqrt{\bar{\alpha}_{t-1}} \left( \frac{x_t - \sqrt{1 - \bar{\alpha}_t} \hat{\epsilon}}{\sqrt{\bar{\alpha}_t}} \right) + \sqrt{1 - \bar{\alpha}_{t-1}} \hat{\epsilon}$
		\EndFor
		\State \textbf{return} $x_0$
	\end{algorithmic}
\end{algorithm}

As described in Algorithm~\ref{alg:sampling}, DDIM Sampling with Deepfake-Resistant Guidance begins by sampling $x_T$ from a standard Gaussian distribution and proceeds through $T$ iterative denoising steps to gradually transform $x_T$ into $x_0$. 
At each timestep $t$, we first construct the facial condition $x_c$ by scaling the cover image $x_{\text{co}}$ with the noise coefficient $\sqrt{1 - \bar{\alpha}_{t-1}}$.
The noised image $x_t$, facial condition $x_c$, timestep $t$, and watermark $w$ are then fed into the diffusion encoder $\mathcal{E}_\theta$ to predict the watermarked image $\hat{x}_0$. When the boolean variable $cond$ is set to true, the Deepfake-resistant guidance can take effect, which subsequently influences the $t$-step noised image $x_t$. 
Specifically, the predict image $\hat{x}_0$ is processed by the Deepfake model $\mathcal{D}_f$ to produce the forged image. Subsequently, the watermark decoder $\mathcal{D}_\theta$ extracts the watermark from this forged image. We then compute the sum of the log‑probabilities across all positions in the extracted watermark sequence. This scalar value is backpropagated with respect to the noised image $x_t$, yielding a gradient that guides the transition from $x_t$ to $x_{t-1}$.
After $T$-step iterations, we will finally obtain the corresponding watermarked image $x_0$ with enhanced robustness. 

In terms of watermark extraction, unlike watermark embedding that is achieved through the multi-step sampling process of conditional diffusion model, watermark extraction is independent of this process. It requires only a single-step decoding operation with the watermark decoder $\mathcal{D}_\theta$ for source tracing.

\subsection{Cross Information Fusion Module}
\label{subsec:CIF}

\begin{figure}[!t]
	\centering
	\includegraphics[width=\columnwidth]{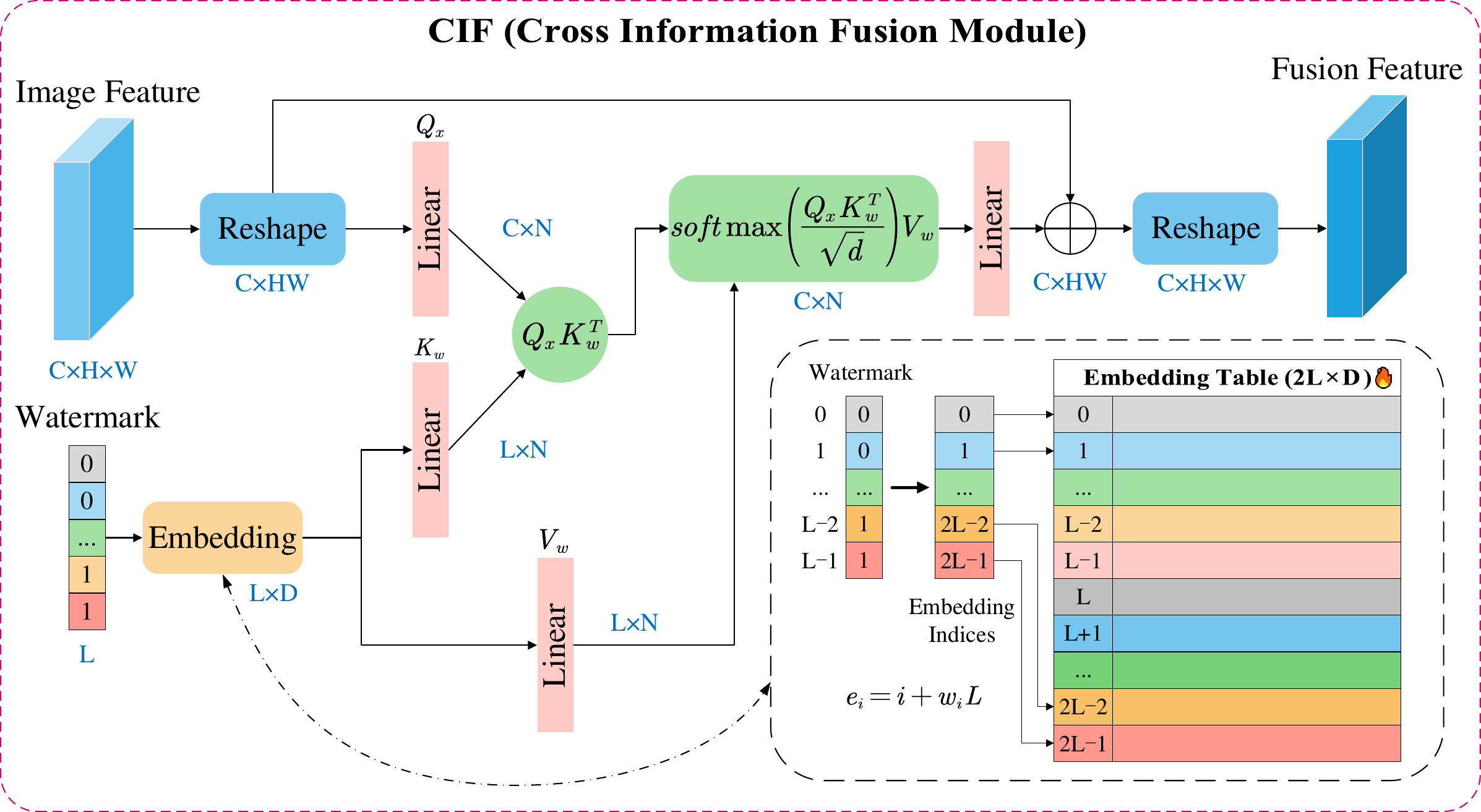}
	\caption{Cross information fusion module. This module combines a learnable embedding table with a cross-attention mechanism.}
	\label{fig:CIF}
\end{figure}

As illustrated in Fig. \ref{fig:CIF}, considering the dimensional discrepancy between the binary watermark \(w \in \{0, 1\}^{L}\) and the image features \(X \in \mathbb{R}^{C \times H \times W}\), which poses challenges for feature fusion, we propose a cross information fusion (CIF) module to address the fusion of image and watermark features in the intermediate layers of diffusion encoder.
The cross information fusion module involves a learnable embedding table \(E \in \mathbb{R}^{2L \times D}\) that establishes continuous embeddings for each bit value in the binary watermark sequence. We develop a simple equation that generates unique embedding index through positional-binary synthesis:
\begin{equation}
	e_i = i + w_i L  \label{eq:embedding_table}
\end{equation} 
where $i$ is the watermark index, $w_i$ is the bit value at position $i$, $L$ is the watermark length, and $e_i$ is the embedding index.

The embedding table can transform the 1D binary sequence into a 2D feature representation \( E_w = E[e_0,\ldots,e_{L-1}] \) 
through the embedding indices, where \( E_w \in \mathbb{R}^{L \times D} \), and \( D \) denotes the dimensionality of watermark features. 
Each watermark bit \( w_i \) can be converted to an embedding index \( e_i \) through the Equation \ref{eq:embedding_table}, ensuring that each bit in the watermark sequence derives a unique feature representation from the embedding table. 

The Equation \ref{eq:embedding_table} builds a bridge between the binary watermark and the embedding indices. The watermark can be deduced reversely from the embedding indices by simple transformation:
\begin{equation}
	w_i = \frac{e_i - i}{L} \label{eq:watermark}
\end{equation} 

To facilitate integration, the image features are correspondingly processed through spatial flattening and dimensional reduction to align with the 2D structure of \( E_w \).
The cross information fusion module then performs feature fusion via cross-attention, where the image and watermark features are first projected through separate linear layers:

\begin{equation}
	Q_x = W_q X_\text{f}, \quad K_w = W_k E_w , \quad V_w = W_v E_w
\end{equation}
where \( W_q \), \( W_k \) and \( W_v \) are learnable projection matrices, and \( X_{\text{f}} \) denotes flattened image features \( X \).

In the fusion process, cross-attention is applied with image features as queries and watermark features as keys and values:

\begin{equation}
	X_{\text{att}} = \text{softmax}\left(\frac{Q_xK_w^T}{\sqrt{d}}\right)V_w
\end{equation}
where \( d \) represents the dimension-normalized scaling factor. A residual connection is then employed as follows:

\begin{equation} 
	X_f^{\text{out}} = X_{f} + X_{\text{att}}
\end{equation}

This design preserves the original image features through the residual connection while effectively integrating the semantic features of the watermark.
Finally, we reshape $X_f^{\text{out}}$ back into the spatial domain to obtain the fused feature map $X_{\text{out}}$, ensuring alignment with the spatial structure of the input image feature.

To provide a clearer understanding of the proposed cross information fusion procedure, we summarize its implementation details in Algorithm~\ref{alg:cif}.

\begin{algorithm}[H]
		\caption{Cross Information Fusion Module}\label{alg:cif}
		\begin{algorithmic}[1]
			\State \textbf{Input:} image feature $X \in \mathbb{R}^{C \times H \times W}$, binary watermark $w \in \{0,1\}^L$, embedding table $E \in \mathbb{R}^{2L \times D}$
			\State \textbf{Output:} fused feature $X_{\text{out}} \in \mathbb{R}^{C \times H \times W}$
			\For{$i = 0$ to $L-1$}
			\State $e_i \leftarrow i + w_i L$ \Comment{positional-binary index, Eq.~(6)}
			\EndFor
			\State $E_w \leftarrow E[\;e_0,\ldots,e_{L-1}\;] \in \mathbb{R}^{L \times D}$ \Comment{embedding lookup}
			\State $X_f \leftarrow \mathrm{Flatten}(X) \in \mathbb{R}^{C \times HW}$
			\State $Q_x \leftarrow W_q X_f$; \quad $K_w \leftarrow W_k E_w$; \quad $V_w \leftarrow W_v E_w$ \Comment{Eq.~(8)}
			\State $X_{\text{att}} \leftarrow \text{softmax}\left(\frac{Q_xK_w^T}{\sqrt{d}}\right)V_w$ \Comment{cross-attention, Eq.~(9)}
			\State $X_f^{\text{out}} \leftarrow X_f + X_{\text{att}}$ \Comment{residual fusion, Eq.~(10)}
			\State $X_{\text{out}} \leftarrow \mathrm{Reshape}(X_f^{\text{out}}) \in \mathbb{R}^{C \times H \times W}$
			\State \textbf{return} $X_{\text{out}}$
		\end{algorithmic}
\end{algorithm}

\subsection{Loss Functions}
\label{sec:loss}
The loss function includes two optimization objectives: the imperceptibility of watermark embedding and the accuracy of watermark extraction.

To address the optimization objective of imperceptibility in watermark embedding, we adopt the mean squared error (MSE) loss at first, which is also the basic loss function in diffusion models:
\begin{equation}
	\mathcal{L}_{\text{mse}} = \|x_0 - \hat{x}_0\|_2^2
\end{equation}

Subsequently, we incorporate the Learned Perceptual Image Patch Similarity (LPIPS) \cite{8578166} loss  to enhance the quality of the watermarked image, ensuring that it is visually indistinguishable from the original image. 
Unlike PSNR and SSIM, which focus on low-level pixel differences, LPIPS captures high-level semantic similarity through deep features from pretrained neural networks, making it more aligned with human perception. Specifically, the quality loss is defined as:
\begin{equation}
	\mathcal{L}_{\text{quality}} = \mathcal{L}_{\text{mse}} + \alpha \mathcal{L}_\text{lpips}(\hat{x}_0, x_0) \label{eq:quality}
\end{equation}
where \( \alpha \) is a loss weight constant.

To address the optimization objective of the accuracy of watermark extraction, we employ the cross-entropy loss. 
Note Equation \ref{eq:embedding_table} that combines position and bit value to generate unique embedding index. 
There are a total of \(2L\) indices and each embedding index points to the unique feature representation of the corresponding bit in binary watermark. In the experiment, we found that when using the mean squared error (MSE) or binary cross-entropy (BCE) loss to direct compare the binary watermark in the case of 256-size image, the DiffMark fails to converge. 
Thus, we try to convert the direct comparison of binary watermarks into the comparison of the corresponding embedding indices according to Equation \ref{eq:embedding_table}. We regard \(2L\) indices as \(2L\) categories and employ the cross-entropy (CE) loss for watermark extraction:
\begin{equation}
	\mathcal{L}_{\text{recovery}} = \mathcal{L}_\text{ce}(\hat{w}, w)
\end{equation}
where \( w \in \{0, 1\}^{L} \) denotes the original binary watermark. The mapping function \( \phi \), defined in Equation \ref{eq:embedding_table}, converts the binary watermark into its corresponding embedding indices \(\phi(w) \in \{0,1,\ldots,2L-1\}^L\), which serve as classification labels. Additionally, \(\hat{w} \in \mathbb{R}^{L \times 2L}\) represents the predicted probabilities for the classes defined by these embedding indices. Specifically:
\begin{equation}
	\mathcal{L}_\text{ce}(\hat{w}, w) = -\frac{1}{L}\sum_{i=1}^L \sum_{k=1}^{2L} \phi(w_{i,k}) \log(\hat{w}_{i,k})
\end{equation}
where \(\phi(w_{i,k})\) denotes the ground truth one-hot encoding, and \(\hat{w}_{i,k}\) represents the predicted probability for class \(k\) at the \(i\)-th position of watermark sequence.

The watermark decoder outputs the probability value of the embedding table indices, not directly the binary watermark. By comparing the embedding indices, our DiffMark could converge at the $256\times256$ image resolution.
We suppose that watermark embedding depends on the embedding indices, it may be more advantageous during training to compare the consistency of corresponding indices rather than the binary watermark itself. 
Due to the reversible transformation of the Equation \ref{eq:embedding_table}, the binary watermark and its corresponding embedding table indices are equivalent.
The original watermark can be recovered from embedding indices using Equation \ref{eq:watermark}.

To summarize, the total loss for the both optimization objectives can be formulated by:
\begin{equation}
	\mathcal{L}_{\text{total}} = \mathcal{L}_{\text{quality}} + \beta \mathcal{L}_{\text{recovery}} \label{eq:total}
\end{equation}
where the loss weight \( \beta \) controls the trade-off between image quality and watermark recovery.

\section{EXPERIMENTS}
\subsection{Experimental Settings}
\subsubsection{Datasets}
In our experiments, two public face datasets, namely CelebA-HQ \cite{karras2017progressive} and LFW \cite{huang2008labeled}, are adopted. The CelebA-HQ dataset contains 30,000 high-resolution $1024\times1024$ facial images. We adopted the official split, where 24,183, 2,993 and 2,824 facial images are used for training, validation, and testing, respectively. The LFW dataset contains 13,233 facial images, each with a resolution of $250\times250$. We randomly select 2,000 images of different individuals to evaluate the generalizability. All the images from both datasets are resized to two resolutions of $128\times128$ and $256\times256$ to accommodate computational resource constraints.
\subsubsection{Implementation Details}
Our DiffMark is implemented by PyTorch \cite{NEURIPS2019_bdbca288} and executed on NVIDIA RTX 3090Ti. 
The pre-trained frozen autoencoder in the training phase is VQGAN \cite{esser2021taming}. 
Our DiffMark is trained on CelebA-HQ dataset for 151.2k steps with a batch size of 16, which is equivalent to 100 epochs. We use the AdamW optimizer \cite{loshchilov2017decoupled} with a learning rate of 1e-4. In Equation~\ref{eq:quality}, the weight $\alpha$ is set to 0.1. To balance the visual quality and watermark robustness, we initialize the weight $\beta$ to 10 in Equation~\ref{eq:total}, then reduce it to 1 after 5k and 10k optimization steps for the $128\times128$ and $256\times256$ resolutions respectively. 
For the embedding dimensionality of the embedding table, we set it to 256 for $128\times128$ images and 1024 for $256\times256$ images respectively. 
To conserve memory under limited computation, we precompute the noised-watermarked image gap with detached computation graphs, and then add it back to preserve gradient flow.
Considering the long sampling time with 1000 steps in the original diffusion model, we reduced the training diffusion steps to 100 and used the DDIM sampler \cite{song2020denoising} with 10 steps during sampling.
\subsubsection{Comparison}
The contrastive methods encompass several deep watermarking methods, including MBRS \cite{jia2021mbrs}, CIN \cite{ma2022towards}, ARWGAN \cite{huang2023arwgan}, SepMark \cite{wu2023sepmark}, EditGuard \cite{zhang2024editguard} and LampMark \cite{wang2024lampmark}.
For SepMark \cite{wu2023sepmark}, we directly adopted the official pre-trained weights, as the dataset and experimental settings are largely consistent. For the other methods, we used their officially released codes and trained them on the CelebA-HQ dataset for 100 epochs. 
Since our objective was to compare the robustness of binary watermark, we only trained the binary watermark network in EditGuard \cite{zhang2024editguard}.
Furthermore, we standardized the length of watermark to 30 and 128 for facial images with resolutions of $128\times128$ and $256\times256$, respectively. 
We re-construct landmark perceptual watermarks for LampMark \cite{wang2024lampmark}, following its original method, to achieve consistent watermark length with other comparative methods.
To compare the robustness of watermarks against Deepfake manipulations, we selected SimSwap \cite{chen2020simswap}, UniFace \cite{xu2022designing}, CSCS \cite{huang2024identity}, StarGAN \cite{choi2018stargan}, and FSRT \cite{rochow2024fsrt} as typical Deepfake methods, covering three major Deepfake categories: face swapping, face attribute editing and face reenactment.
\subsubsection{Evaluation Metrics}
The evaluation of the image watermarking task includes two aspects: the invisibility of watermark embedding and the robustness of watermark extraction. For the invisibility of watermark embedding, we use three metrics: the average peak signal-to-noise ratio (PSNR), the structural similarity index measure (SSIM), and the learned perceptual image patch similarity (LPIPS). For the robustness of watermark extraction, we use the bit error ratio (BER) in the DiffMark:
\begin{equation}
	\label{ber}
	BER(\hat{w}, w) =  \frac{1}{L} \times  \sum_{i=1}^{L} 1\left( \varphi(\hat{w}_{i}) \neq w_{i} \right) \times 100\%
\end{equation}
where $1\left( \varphi(\hat{w}_{i}) \neq w_{i}  \right)$
is an indicator function that outputs 0 if the extracted watermark bit matchs the embedded watermark bit at the corresponding position; otherwise, it outputs 1. The \( \varphi \) represents the mapping function defined in Equation \ref{eq:watermark}, which converts the extracted embedding indices back to the binary watermark.
\subsection{Intra-Dataset Evaluation}

\subsubsection{Visual Quality}
In the watermark embedding invisibility experiment, we evaluated the average PSNR, SSIM and LPIPS between the original and watermarked image.
These three metrics are used to assess the differences in image quality, structural similarity, and perceptual similarity between the watermarked image and the original image. 
As shown in Table \ref{tab:visual_quality}, our DiffMark maintains the best visual quality at both $128\times128$ and $256\times256$ image resolutions, outperforming existing watermarking methods. 
This indicates that the watermarked images generated by our DiffMark are very similar to the original images, with almost no perceptible visual differences. 
The performance of ARWGAN \cite{huang2023arwgan} substantially deteriorates at $256\times256$ resolution, primarily due to its difficulties in network convergence at slightly higher resolutions.
\begin{table}[!htbp]
	\centering
	\caption{Quantitative Visual Quality Evaluation of the Watermarked Images.}
	\label{tab:visual_quality}
	\resizebox{0.6\columnwidth}{!}{
		\begin{tabular}{lcccccc}
			\toprule
			& \multicolumn{3}{c}{$128\times128$ / 30} & \multicolumn{3}{c}{$256\times256$ / 128} \\
			\cmidrule(lr){2-4} \cmidrule(lr){5-7}
			Methods & PSNR ↑ & SSIM ↑ & LPIPS ↓ & PSNR ↑ & SSIM ↑ & LPIPS ↓ \\
			\midrule
			MBRS \cite{jia2021mbrs} & 35.1897 & 0.9021 & 0.0744 & 36.3383 & 0.8857 & 0.1188 \\
			CIN \cite{ma2022towards} & 39.7044 & 0.9308 & 0.0248 & 37.1549 & 0.8483 & 0.0705 \\
			ARWGAN \cite{huang2023arwgan} & 38.5746 & 0.9733 & 0.0183 & 29.6473 & 0.8271 & 0.2792 \\
			SepMark \cite{wu2023sepmark} & 38.3129 & 0.9599 & 0.0196 & 38.4669 & 0.9339 & 0.0407 \\
			EditGuard \cite{zhang2024editguard} & 37.1664 & 0.9516 & 0.0746 & 36.6557 & 0.8910 & 0.1496 \\
			LampMark \cite{wang2024lampmark} & 40.2231 & 0.9666 & 0.0293 & 39.5722 & 0.9515 & 0.0715 \\
			Ours & \textbf{41.2869} & \textbf{0.9776} & \textbf{0.0090} & \textbf{41.9572} & \textbf{0.9769} & \textbf{0.0116} \\
			\bottomrule
		\end{tabular}
	}
\end{table}

\subsubsection{Robustness}

In the watermark robustness experiment, we used the average BER as the evaluation metric. As the accuracy of watermark extraction is inversely proportional to the Bit Error Rate (BER), a lower average BER under various distortions suggests better robustness of the watermark.

\begin{table}[!htbp]
	\centering
	\caption{Quantitative Comparison on CelebA-HQ regarding
		Bit Error Rate (BER) of the Watermarks
		under Benign Distortions.}
	\label{tab:robustness_benign}
	\resizebox{\textwidth}{!}{
		\begin{tabular}{lcccccccccccccc}
			\toprule
			& \multicolumn{2}{c}{MBRS \cite{jia2021mbrs}} & \multicolumn{2}{c}{CIN \cite{ma2022towards}} & \multicolumn{2}{c}{ARWGAN \cite{huang2023arwgan}} & \multicolumn{2}{c}{SepMark \cite{wu2023sepmark}} & \multicolumn{2}{c}{EditGuard \cite{zhang2024editguard}} & \multicolumn{2}{c}{LampMark \cite{wang2024lampmark}} & \multicolumn{2}{c}{Ours} \\
			\cmidrule(lr){2-3} \cmidrule(lr){4-5} \cmidrule(lr){6-7} \cmidrule(lr){8-9} \cmidrule(lr){10-11} \cmidrule(lr){12-13} \cmidrule(lr){14-15}
			Distortion & 128   & 256   & 128   & 256   & 128   & 256   & 128   & 256   & 128   & 256   & 128   & 256  & 128   & 256 \\
			\midrule
			Identity  & 0.00\% & 0.00\% & 0.00\% & 0.00\% & 0.00\% & 9.86\% & 0.00\% & 0.00\% & 0.09\% & 0.12\% & 0.00\% & 0.00\% & 0.00\% & 0.00\% \\
			Resize(p=0.8) & 10.72\% & 13.58\% & 24.97\% & 36.21\% & 0.02\% & 10.10\% & 23.81\% & 3.40\% & 0.60\% & 0.28\% & 14.00\% & 13.82\% & 0.02\% & 0.01\% \\
			Dropout(p=0.6) & 0.71\% & 8.64\% & 0.00\% & 0.00\% & 0.69\% & 12.44\% & 0.35\% & 0.28\% & 1.02\% & 0.97\% & 1.91\% & 2.43\% & 0.31\% & 0.60\% \\
			GaussianNoise(s=0.1) & 0.07\% & 0.08\% & 0.00\% & 0.00\% & 20.31\% & 11.65\% & 0.76\% & 0.06\% & 1.67\% & 1.02\% & 9.57\% & 7.89\% & 1.35\% & 2.37\% \\
			SaltPepper(p=0.1) & 12.49\% & 12.40\% & 0.00\% & 0.00\% & 0.00\% & 9.84\% & 0.02\% & 0.00\% & 0.09\% & 0.12\% & 24.09\% & 23.82\% & 0.09\% & 0.48\% \\
			GaussianBlur(k=5,s=5) & 4.57\% & 10.14\% & 6.89\% & 0.39\% & 6.97\% & 22.72\% & 0.41\% & 0.04\% & 1.53\% & 3.37\% & 1.82\% & 0.59\% & 0.00\% & 0.00\% \\
			MedianBlur(k=5) & 1.11\% & 5.55\% & 0.81\% & 0.65\% & 3.07\% & 18.82\% & 0.20\% & 0.03\% & 1.62\% & 3.85\% & 1.72\% & 0.63\% & 0.00\% & 0.00\% \\
			Brightness(f=0.5) & 0.04\% & 0.15\% & 0.00\% & 0.00\% & 0.09\% & 12.01\% & 0.00\% & 0.00\% & 4.53\% & 2.58\% & 0.29\% & 0.32\% & 0.37\% & 0.64\% \\
			Contrast(f=0.5) & 0.02\% & 0.13\% & 0.00\% & 0.00\% & 0.10\% & 11.56\% & 0.00\% & 0.00\% & 0.51\% & 0.32\% & 0.25\% & 0.32\% & 0.34\% & 0.73\% \\
			Saturation(f=0.5) & 0.00\% & 0.00\% & 0.00\% & 0.00\% & 0.01\% & 11.15\% & 0.00\% & 0.00\% & 0.10\% & 0.14\% & 0.00\% & 0.00\% & 0.00\% & 0.00\% \\
			Hue(f=0.1) & 0.00\% & 0.00\% & 0.00\% & 0.00\% & 2.56\% & 15.68\% & 0.54\% & 0.00\% & 0.34\% & 0.17\% & 0.00\% & 0.00\% & 0.27\% & 0.42\% \\
			JpegTest(Q=50) & 0.00\% & 0.01\% & 5.25\% & 8.90\% & 16.47\% & 15.34\% & 1.22\% & 0.10\% & 1.42\% & 3.50\% & 0.69\% & 1.83\% & 0.85\% & 1.81\% \\
			\midrule
			Average & 2.48\% & 4.22\% & 3.16\% & 3.85\% & 4.19\% & 13.43\% & 2.28\% & \textbf{0.33\%} & 1.13\% & 1.37\% & 4.53\% & 4.31\% & \textbf{0.30\%} & 0.59\% \\
			\bottomrule
		\end{tabular}
	}
	
\end{table}

In Table \ref{tab:robustness_benign}, we evaluated the method using various distortions such as \{Identity, Resize, Dropout, GaussianNoise, SaltPepper, GaussianBlur, MedianBlur, Brightness, Contrast, Saturation, Hue, JpegTest\}. It can be seen that our DiffMark, along with the SepMark, achieves the lowest average watermark BER at $128\times128$ and $256\times256$ resolutions respectively. However, since our DiffMark generates watermarked images by denoising based on the standard Gaussian distribution, the BER slightly increases when facing Gaussian noise, indicating some sensitivity to this type of distortion. 
It is observed that the compared watermarking methods consistently exhibit vulnerability to specific distortions like resize and salt-and-pepper noise, likely due to insufficient consideration of these distortions in their network architecture or noise layer design.
Notably, during training, we only incorporated a frozen-parameter autoencoder and did not include these common distortions in our end-to-end training framework. Despite this, our DiffMark still maintains strong robustness against these distortions, which exceeds our expectations.

\begin{table}[!htbp]
	\centering
	\caption{Quantitative Comparison on CelebA-HQ regarding the Bit Error Rate (BER) of the Watermarks under various Deepfake manipulations.}
	\label{tab:robustness_deepfake}
	\resizebox{\textwidth}{!}{
		\begin{tabular}{lcccccccccccccc}
			\toprule
			& \multicolumn{2}{c}{MBRS \cite{jia2021mbrs}} & \multicolumn{2}{c}{CIN \cite{ma2022towards}} & \multicolumn{2}{c}{ARWGAN \cite{huang2023arwgan}} & \multicolumn{2}{c}{SepMark \cite{wu2023sepmark}} & \multicolumn{2}{c}{EditGuard \cite{zhang2024editguard}} & \multicolumn{2}{c}{LampMark \cite{wang2024lampmark}} & \multicolumn{2}{c}{Ours} \\
			\cmidrule(lr){2-3} \cmidrule(lr){4-5} \cmidrule(lr){6-7} \cmidrule(lr){8-9} \cmidrule(lr){10-11} \cmidrule(lr){12-13} \cmidrule(lr){12-13} \cmidrule(lr){14-15}
			Distortion & 128   & 256   & 128   & 256   & 128   & 256   & 128   & 256   & 128   & 256   & 128   & 256   & 128   & 256 \\
			\midrule
			SimSwap \cite{chen2020simswap}   & 24.60\% & 27.09\% & 40.08\% & 31.07\% & 46.60\% & 41.57\% & 20.02\% & 11.95\% & 45.32\% & 47.99\% & 16.53\% & 15.11\% & 5.58\%  & 5.96\%  \\
			UniFace \cite{xu2022designing}  & 0.48\%  & 26.57\% & 11.80\% & 48.27\% & 26.28\% & 42.69\% & 0.34\%  & 31.93\% & 9.17\%  & 49.41\% & 6.28\% & 29.85\% & 0.01\%  & 2.20\%  \\
			CSCS \cite{huang2024identity}     & 10.10\%  & 10.33\% & 0.29\%  & 0.79\%  & 6.29\%  & 14.82\% & 0.68\%  & 2.35\%  & 0.99\%  & 1.61\% & 2.30\% & 0.63\% & 0.13\%  & 0.56\%  \\
			StarGAN \cite{choi2018stargan}  & 5.49\%  & 17.26\% & 56.93\% & 38.56\% & 36.78\% & 32.35\% & 0.11\%  & 0.01\%  & 7.62\%  & 2.12\% & 7.30\% & 4.05\% & 4.66\%  & 3.82\%  \\
			FSRT \cite{rochow2024fsrt}     & 2.40\% & 20.92\% & 3.20\%  & 35.22\% & 4.34\%  & 35.32\% & 0.78\% & 9.21\%  & 5.77\%  & 31.17\% & 2.93\% & 18.08\% & 0.12\%  & 4.05\%  \\
			VQGAN \cite{esser2021taming}    & 0.05\%  & 0.73\%  & 39.60\% & 24.90\% & 35.06\% & 34.86\% & 1.28\%  & 0.10\%  & 7.49\%  & 6.38\% & 10.22\% & 10.75\% & 0.02\%  & 0.02\%  \\
			\midrule
			Average   & 7.19\% & 17.15\% & 25.32\% & 29.80\% & 25.89\% & 33.60\% & 3.87\% & 9.26\%  & 12.73\% & 23.11\% & 7.59\% & 13.08\% & \textbf{1.75\%}  & \textbf{2.77\%}  \\
			\bottomrule
		\end{tabular}
	}
	
\end{table}

In Table \ref{tab:robustness_deepfake}, we conducted the evaluation of BER under representative Deepfake manipulations such as \{SimSwap, UniFace, CSCS, StarGAN, FSRT\}.
In this study, the VQGAN serves as an autoencoder to simulate Deepfake manipulations in our training framework and we include it in Table \ref{tab:robustness_deepfake} and Table \ref{tab:cross_dataset} for comparative reference.
The experimental results demonstrate that our method achieves lower BER against most Deepfake manipulations, outperforming many of comparison methods. 
Notably, SepMark exhibits the lowest BER on StarGAN, which may be due to the targeted optimization of StarGAN within its training framework. 
However, when considering the average BER across both $128\times128$ and $256\times256$ resolutions, our DiffMark achieves the best performance, confirming its generalization capability in handling various Deepfake manipulations.
Furthermore, we observe that even the same Deepfake model can exhibit varying impacts on watermark robustness across different image resolutions, as seen with UniFace \cite{xu2022designing} and FSRT \cite{rochow2024fsrt}. It may be because the Deepfake model tends to bring more distortions to the watermarked image as the image resolution increases.

\subsection{Cross-Dataset Evaluation}

To further evaluate the generalization of DiffMark in cross-dataset settings, we conducted experiments on the LFW dataset at both $128\times128$ and $256\times256$ resolutions. 
\begin{table}[!htbp]
	\centering
	\caption{Quantitative Experiments on LFW Dataset for Visual Quality and Bit Error Rate (BER) of the Watermarks
		under Deepfake Manipulations.}
	\label{tab:cross_dataset}
	\resizebox{\textwidth}{!}{
		\begin{tabular}{lcccccccccccccc}
			\toprule
			& \multicolumn{2}{c}{MBRS \cite{jia2021mbrs}} & \multicolumn{2}{c}{CIN \cite{ma2022towards}} & \multicolumn{2}{c}{ARWGAN \cite{huang2023arwgan}} & \multicolumn{2}{c}{SepMark \cite{wu2023sepmark}} & \multicolumn{2}{c}{EditGuard \cite{zhang2024editguard}} & \multicolumn{2}{c}{LampMark \cite{wang2024lampmark}} & \multicolumn{2}{c}{Ours} \\
			\cmidrule(lr){2-3} \cmidrule(lr){4-5} \cmidrule(lr){6-7} \cmidrule(lr){8-9} \cmidrule(lr){10-11} \cmidrule(lr){12-13} \cmidrule(lr){14-15}
			& 128 & 256 & 128 & 256 & 128 & 256 & 128 & 256 & 128 & 256 & 128 & 256 & 128 & 256 \\
			\midrule
			SimSwap \cite{chen2020simswap} & 26.38\% & 24.23\% & 43.02\% & 27.47\% & 47.74\% & 42.56\% & 25.91\% & 18.07\% & 45.53\% & 47.40\% & 24.95\% & 15.61\% & 9.03\% & 8.35\% \\
			UniFace \cite{xu2022designing} & 0.14\% & 21.16\% & 12.26\% & 47.99\% & 26.90\% & 42.34\% & 0.41\% & 25.52\% & 8.98\% & 48.80\% & 10.17\% & 31.00\% & 0.01\% & 2.07\% \\
			CSCS \cite{huang2024identity} & 5.27\% & 6.53\% & 0.31\% & 0.70\% & 14.62\% & 15.12\% & 1.73\% & 1.06\% & 4.29\% & 2.33\% & 10.32\% & 1.38\% & 0.64\% & 3.82\% \\
			StarGAN \cite{choi2018stargan} & 6.19\% & 17.29\% & 58.15\% & 43.18\% & 40.66\% & 33.29\% & 0.51\% & 0.04\% & 12.94\% & 3.01\% & 17.00\% & 5.49\% & 6.03\% & 5.45\% \\
			FSRT \cite{rochow2024fsrt} & 4.01\% & 19.69\% & 7.84\% & 38.36\% & 11.64\% & 36.83\% & 1.54\% & 14.26\% & 12.90\% & 35.65\% & 15.21\% & 25.46\% & 0.43\% & 9.93\% \\
			VQGAN \cite{esser2021taming} & 0.17\% & 0.13\% & 40.13\% & 18.10\% & 39.82\% & 33.28\% & 1.17\% & 0.16\% & 14.38\% & 5.87\% & 21.53\% & 6.49\% & 0.04\% & 0.01\% \\
			Average & 7.03\% & 14.84\% & 26.95\% & 29.30\% & 30.23\% & 33.90\% & 5.21\% & 9.85\% & 16.50\% & 23.84\% & 16.53\% & 14.24\% & \textbf{2.70\%} & \textbf{4.94\%} \\
			\midrule
			PSNR ↑ & 34.99 & 36.73 & 39.64 & 37.21 & 38.60 & 29.69 & 37.30 & 38.28 & 34.98 & 34.23 & 37.18 & 39.56 & 39.10 & 41.33 \\
			SSIM ↑ & 0.900 & 0.880 & 0.930 & 0.824 & 0.973 & 0.851 & 0.951 & 0.930 & 0.938 & 0.840 & 0.937 & 0.947 & 0.970 & 0.973 \\
			LPIPS ↓ & 0.080 & 0.144 & 0.025 & 0.094 & 0.019 & 0.178 & 0.027 & 0.055 & 0.093 & 0.217 & 0.056 & 0.085 & 0.013 & 0.011 \\
			\bottomrule
		\end{tabular}
	}
	
\end{table}
As shown in Table \ref{tab:cross_dataset}, the experimental results indicate that the performance trends of most watermarking methods are largely consistent with those observed on CelebA-HQ. Specifically, in terms of watermark robustness against Deepfakes, the BER values for most methods are generally slight higher than those on CelebA-HQ. This suggests that most watermarking methods lack generalization ability and consequently remain vulnerable to Deepfake manipulations. 
Regarding watermark imperceptibility, quantitative metrics such as PSNR, SSIM, and LPIPS reveal that the performance of DiffMark on LFW is slightly inferior to its benchmark results on CelebA-HQ. This performance gap can be attributed to the inherent differences in dataset characteristics: while CelebA-HQ dataset comprises high-quality facial images, LFW dataset contains a substantial number of low-resolution and blurred samples. The distributional discrepancy consequently leads to the reduction of watermark invisibility during the diffusion sampling process.

\subsection{Ablation Studies}
\subsubsection{Hyperparameters $\alpha$ and $\beta$}

\begin{table}[htbp]
	\centering
	\caption{Ablation Study on the Hyperparameters $\alpha$ and $\beta$ (Image Size $128\times128$). The Table Shows Image Quality Metrics (PSNR, SSIM, LPIPS) and Watermark Bit Error Rates (BER) under Various Deepfake Manipulations.}
	\label{tab:ablation_ab}
	\resizebox{\textwidth}{!}{
		\begin{tabular}{cc|ccc|ccccccc}
			\toprule
			\multirow{2}{*}{$\alpha$} & \multirow{2}{*}{$\beta$} &
			\multicolumn{3}{c|}{Image Quality Metrics} &
			\multicolumn{7}{c}{Watermark Bit Error Rates (BER) $\downarrow$} \\
			\cmidrule(lr){3-5}\cmidrule(lr){6-12}
			& & PSNR$\uparrow$ & SSIM$\uparrow$ & LPIPS$\downarrow$ &
			SimSwap & UniFace & CSCS & StarGAN & FSRT & VQGAN & Average \\
			\midrule
			0.1  & 1.0           & 37.9254 & 0.9579 & 0.0202 & 2.86\% & 0.00\% & 0.06\% & 2.60\% & 0.01\% & 0.00\% & 0.92\% \\
			0.1  & 0.1           & 48.1395 & 0.9946 & 0.0007 & 50.14\% & 49.93\% & 49.80\% & 50.43\% & 50.35\% & 50.09\% & 50.12\% \\
			0.1  & $1.0\to0.1$   & 41.2869 & 0.9776 & 0.0090 & 5.58\% & 0.01\% & 0.13\% & 4.66\% & 0.12\% & 0.02\% & 1.75\% \\
			1.0  & $1.0\to0.1$   & 40.1124 & 0.9657 & 0.0047 & 9.50\% & 0.01\% & 0.17\% & 10.39\% & 0.66\% & 0.16\% & 3.48\% \\
			0.01 & $1.0\to0.1$   & 41.2220 & 0.9743 & 0.0165 & 2.69\% & 0.00\% & 0.01\% & 2.03\% & 0.07\% & 0.00\% & \textbf{0.80}\% \\
			\bottomrule
		\end{tabular}
	}
\end{table}

In this section, we perform an ablation study on two hyperparameters, $\alpha$ and $\beta$, to evaluate their effects on the watermarking performance. The results in Table~\ref{tab:ablation_ab} highlight the trade-off between watermark robustness and image quality. When $\alpha$ is fixed at 0.1, the model fails to converge with $\beta = 0.1$. Increasing $\beta$ to 1.0 reduces the average bit error rate (BER) but at the cost of lower image quality. Annealing $\beta$ from 1.0 to 0.1 strikes a better balance between robustness and invisibility. Similarly, with $\beta$ annealed from 1.0 to 0.1, a smaller $\alpha$ value further decreases BER, yet also degrades image quality. However, increasing $\alpha$ to 1.0 significantly diminishes watermark robustness without improving image quality. In comparison, setting $\alpha = 0.1$ yields a more desirable outcome in terms of both metrics. These results indicate that improved robustness often comes at the expense of visual fidelity—smaller $\alpha$ and larger $\beta$ could enhance robustness but reduce image quality. The optimal balance is dictated by application priorities.

\subsubsection{Cross Information Fusion Module}

To analyze the impact of the embedding dimension in the Cross Information Fusion (CIF) module, we conduct an ablation study on the dimension of the embedding table.
\begin{table}[h]
	\centering
	\caption{Ablation Study on the Dimension of Embedding Table in the CIF Module (Image Size $128\times128$). We Report Image Quality (PSNR, SSIM, LPIPS) and Watermark Bit Error Rates (BER) under Various Deepfake Manipulations.}
	\label{tab:ablation_cif}
	\resizebox{\textwidth}{!}{
		\begin{tabular}{c|ccc|ccccccc}
			\toprule 
			\multirow{2}{*}{Embedding Dim} & \multicolumn{3}{c|}{Image Quality Metrics} & \multicolumn{7}{c}{Watermark Bit Error Rates (BER) $\downarrow$} \\
			\cmidrule{2-11}
			& PSNR$\uparrow$ & SSIM$\uparrow$ & LPIPS$\downarrow$ & SimSwap & UniFace & CSCS & StarGAN & FSRT & VQGAN & Average \\
			\midrule 
			128   & 41.1802 & 0.9754 & 0.0104 & 6.71\% & 0.02\% & 0.31\% & 5.89\% & 0.12\% & 0.03\% & 2.18\% \\
			256   & 41.2869 & 0.9776 & 0.0090 & 5.58\% & 0.01\% & 0.13\% & 4.66\% & 0.12\% & 0.02\% & 1.75\% \\
			512   & 40.8905 & 0.9744 & 0.0104 & 5.09\% & 0.00\% & 0.07\% & 3.70\% & 0.18\% & 0.02\% & 1.51\% \\
			1024  & 40.7391 & 0.9749 & 0.0107 & 4.39\% & 0.00\% & 0.04\% & 3.18\% & 0.16\% & 0.02\% & \textbf{1.30\%} \\
			1536  & 40.3886 & 0.9716 & 0.0118 & 5.16\% & 0.00\% & 0.03\% & 3.53\% & 0.20\% & 0.02\% & 1.49\% \\
			\bottomrule 
		\end{tabular}
	}
\end{table}

Table~\ref{tab:ablation_cif} shows that as the embedding dimension increases, the average bit error rate (BER) generally decreases, indicating improved robustness against Deepfake manipulations. However, this improvement plateaus beyond 1024 dimensions, as seen by the rise in BER at 1536 dimensions. This suggests that while increasing the embedding dimension initially enhances robustness, the gains diminish beyond a certain point, and further increases contribute little to additional improvement in watermark robustness. Moreover, the enhanced robustness is often associated with a decrease in image quality, as reflected by lower PSNR and SSIM values and higher LPIPS values.

\subsection{Deepfake-Resistant Guidance}
In this section, we investigate whether incorporating Deepfake-resistant guidance during DDIM sampling enhances watermark robustness against Deepfake manipulations. The gradient scale $s$ in Algorithm~\ref{alg:sampling} is set to 1k. We only incorporate SimSwap \cite{chen2020simswap} as the specific Deepfake model in Deepfake-resistant guidance and test the robustness of watermark across various Deepfake models such as \{SimSwap, UniFace, CSCS, StarGAN, FSRT, VQGAN\}.

\begin{table}[ht]
	\centering
	\caption{Quantitative Experiments of the Deepfake-Resistant Guidance in DDIM Sampling on CelebA-HQ and LFW Dataset.}
	\label{tab:guidance}
	\resizebox{0.6\textwidth}{!}{
		\begin{tabular}{lcccccccc}
			\toprule
			& \multicolumn{4}{c}{CelebA-HQ} & \multicolumn{4}{c}{LFW} \\
			\cmidrule(lr){2-5} \cmidrule(lr){6-9}
			& \multicolumn{2}{c}{$128\times128$} & \multicolumn{2}{c}{$256\times256$} & \multicolumn{2}{c}{$128\times128$} & \multicolumn{2}{c}{$256\times256$}\\
			\cmidrule(lr){2-3} \cmidrule(lr){4-5} \cmidrule(lr){6-7} \cmidrule(lr){8-9}
			Distortion & w/o & w/ & w/o & w/ & w/o & w/ & w/o & w/ \\
			\midrule
			SimSwap \cite{chen2020simswap} & 5.58\% & 1.71\% & 5.96\% & 1.16\% & 9.03\% & 4.61\% & 8.35\% & 1.30\% \\
			UniFace \cite{xu2022designing} & 0.01\% & 0.01\% & 2.20\% & 1.94\% & 0.01\% & 0.00\% & 2.07\% & 1.74\% \\
			CSCS \cite{huang2024identity} & 0.13\% & 0.10\% & 0.56\% & 0.45\% & 0.64\% & 0.56\% & 3.82\% & 2.88\% \\
			StarGAN \cite{choi2018stargan} & 4.66\% & 3.49\% & 3.82\% & 2.60\% & 6.03\% & 4.46\% & 5.45\% & 3.70\% \\
			FSRT \cite{rochow2024fsrt} & 0.12\% & 0.10\% & 4.05\% & 3.43\% & 0.43\% & 0.30\% & 9.93\% & 7.89\% \\
			VQGAN \cite{esser2021taming} & 0.02\% & 0.02\% & 0.02\% & 0.01\% & 0.04\% & 0.02\% & 0.01\% & 0.01\% \\
			Average & 1.75\% & 0.91\% & 2.77\% & 1.60\% & 2.70\% & 1.66\% & 4.94\% & 2.92\% \\
			\midrule
			PSNR $\uparrow$ & 41.29 & 40.96 & 41.96 & 41.03 & 39.10 & 38.82 & 41.33 & 40.14 \\
			SSIM $\uparrow$ & 0.978 & 0.975 & 0.977 & 0.968 & 0.970 & 0.966 & 0.973 & 0.960 \\
			\bottomrule
		\end{tabular}
	}
\end{table}

As presented in Table \ref{tab:guidance}, which reports the bit error rate (BER) for watermark extraction, we observe three findings: 
First, incorporating Deepfake-resistant guidance during DDIM sampling consistently results in a lower BER compared to that achieved without Deepfake-resistant guidance.
Second, the robustness of watermark against SimSwap shows the most significant improvement, while the robustness against other Deepfake models also improves to varying degrees.
Third, although the Deepfake-resistant guidance improves the robustness of the watermark against Deepfake manipulations, this may lead to a little decrease in visual quality of the watermarked facial image.
The experimental results indicate that the Deepfake-resistant guidance can be used as a training-free enhancement module during the diffusion sampling process, it can guide the sampling process to generate more robust watermark images against Deepfake manipulations.

\subsection{Visualization Result}

\begin{figure*}[!t]
	\centering
	\includegraphics[width=\textwidth]{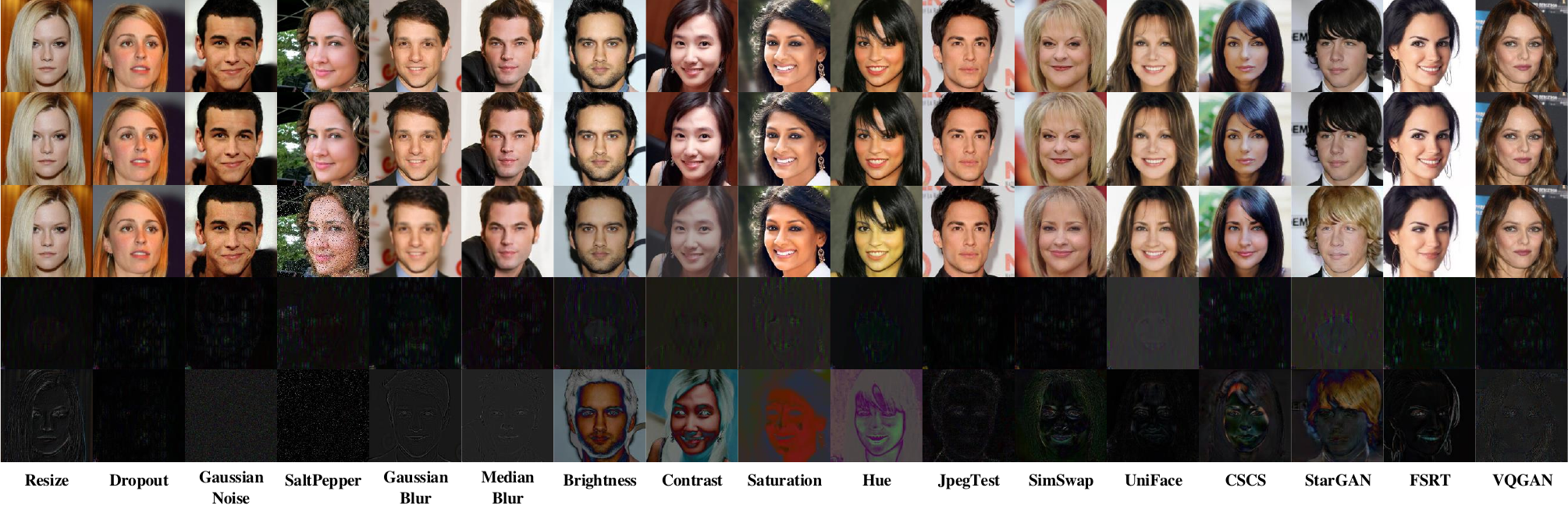}
	\caption{
		The visual quality of facial images under various typical distortions. 
		The rows from top to bottom display: 
		(a) the original cover image \( x_\text{co} \), 
		(b) the watermarked image \( x_\text{wm} \), 
		(c) the distorted image \( x_\text{dt} \), 
		(d) the normalized residual signal between \( x_\text{wm} \) and \( x_\text{co} \), and 
		(e) the normalized residual signal between \( x_\text{dt} \) and \( x_\text{wm} \). 
		Each column represents a distinct distortion type. 
		All images have a size of $256\times256$ pixels.
	}
	\label{fig:vision}
\end{figure*}

The sampled images are shown in Fig. \ref{fig:vision}, with the five rows from top to bottom representing the original image \( x_\text{co} \), the watermarked image \( x_\text{wm} \), the distorted image \( x_\text{dt} \), the residual signal of \( \left| \mathcal{N}(x_\text{wm} - x_\text{co}) - 0.5 \right|\) and \( \left| \mathcal{N}(x_\text{dt} - x_\text{wm}) - 0.5 \right|\), where \( \mathcal{N}(x) = (x - min(x))/(max (x) - min(x)) \). The first 11 columns display the effects of benign distortions, while the remaining columns show the effects of Deepfake manipulations. For simplicity, the original image required for face swapping and reenactment, as well as the specific attributes needed for attribute editing, are omitted. The last column shows the image reconstruction by the VQGAN \cite{esser2021taming} autoencoder. It can be observed that the watermark is embedded into the facial image in an invisible manner, without affecting the visual quality of the image.

\subsection{limitations}

\begin{table}[t]
	\centering
	\caption{Quantitative Experiments on Average Watermark Embedding and Extraction Times, as well as Peak Memory Usage during the Inference Phase, per $128\times128$ Image.}
	\label{tab:time_memory}
	\resizebox{0.6\textwidth}{!}{
		\begin{tabular}{lcccc}
			\toprule
			Method & Embed Time (s) & Extract Time (s) & Peak Mem (GB) \\
			\midrule
			MBRS \cite{jia2021mbrs} & 0.0072 & 0.0065 & 0.1782  \\
			CIN \cite{ma2022towards} & 0.0182 & 0.0171 & 0.2049  \\
			ARWGAN \cite{huang2023arwgan} & 0.0024 & 0.0006 & 0.1916  \\
			SepMark \cite{wu2023sepmark} & 0.0109 & 0.0168 & 0.6150  \\
			EditGuard \cite{zhang2024editguard} & 0.0105 & 0.0084 & 0.1101  \\
			LampMark \cite{wang2024lampmark} & 0.0040 & 0.0039 & 0.1083 \\
			Ours & 0.1320 & 0.0042 & 0.1902 \\
			Ours (guidance) & 0.7502 & 0.0048 & 0.7078 \\
			\bottomrule
		\end{tabular}
	}
\end{table}

In this section, we evaluate and analyze the computational efficiency and memory usage.
As shown in Table~\ref{tab:time_memory}, the watermark extraction time of our method is comparable to that of the baseline methods, as all involve a one-step extraction process. 
However, due to the multi-step nature of the diffusion mechanism, watermark embedding naturally takes longer compared to the baseline methods, which use a one-step embedding process. 
Additionally, incorporating Deepfake-resistant guidance during the sampling process further increases the time due to the extra computations involving the Deepfake model. Nevertheless, the embedding time of 0.7502 seconds remains within an acceptable range for practical use.
In terms of GPU memory usage, our method is similar to others when no guidance is applied. However, enabling Deepfake-resistant guidance increases memory usage, as the Deepfake model must remain loaded throughout the sampling process.
Given the advantages of our method, including the enhanced watermark robustness and invisibility through the diffusion mechanism, we will focus on improving computational efficiency in future research. Despite these limitations, the primary goal of this study is to advance the application of diffusion models in Deepfake proactive forensics, with the hope of providing new insights and approaches for future development.

\section{CONClUSION}

In this work, we propose DiffMark, a diffusion-based robust watermarking framework that constructs facial image and watermark as conditions to guide the diffusion sampling process to progressively denoise and generate watermarked image.
We design a cross information fusion module for the fusion of image features and watermark. To enhance the robustness of the watermark against Deepfake manipulations, we integrate a pre-trained frozen autoencoder during training phase and introduce Deepfake-resistant guidance during sampling phase. Experimental results demonstrate that DiffMark achieves high watermark invisibility and robustness.
Although DiffMark provides traceability and copyright protection for facial images against Deepfake manipulations, its malicious use may raise ethical concerns, particularly regarding privacy violations when applied without consent. Future work could explore the privacy-preserving techniques such as differential privacy. Moreover, malicious actors involved in Deepfake distribution often tend to remove the watermarks in facial images, thereby compromising the effectiveness of watermarks. It is important to develop more robust watermark to address such risk. Lastly, further functionality could be developed. This research introduces Deepfake-resistant guidance to improve watermark traceability, and it may also be possible to leverage adversarial gradient guidance during the diffusion sampling phase to enhance the performance of Deepfake detectors or disrupt the face forgery effects of Deepfake models.

%


  \bibliographystyle{elsarticle-num-names} 
  \bibliography{cas-refs}


\end{document}